\documentclass[runningheads]{llncs}
\usepackage[misc]{ifsym}
\usepackage[T1]{fontenc}
\usepackage{graphicx}
\usepackage{booktabs}

\usepackage{mwe}
\usepackage{times}  
\usepackage{helvet}  
\usepackage{courier}  
\usepackage[hyphens]{url}  
\usepackage{graphicx} 
\usepackage{caption} 
\graphicspath{{images/}}
\usepackage{graphicx}
\usepackage{caption}
\usepackage{soul}
\usepackage{subcaption}

\usepackage[misc]{ifsym}
\usepackage{fancyhdr}
\usepackage{xcolor} 
\usepackage{ifthen}
\usepackage{etoolbox}
\usepackage{algorithm}
\usepackage{algorithmic}
\usepackage{amssymb}
\usepackage{xcolor}
\usepackage{amsmath}
\usepackage{newfloat}
\usepackage{listings}
\DeclareCaptionStyle{ruled}{labelfont=normalfont,labelsep=colon,strut=off} 
\floatstyle{ruled}
\newfloat{listing}{tb}{lst}{}
\floatname{listing}{Listing}

\titlerunning{Federated Learning with Local Update Approximation and Rectification}

\titlerunning{\date{}}  

\authorrunning{Chutian Jiang et al.} 
\institute{}

\title{FedAR: Addressing Client Unavailability in Federated Learning with Local Update Approximation and Rectification}

\author{Chutian Jiang \and Hansong Zhou \and
Xiaonan Zhang \textsuperscript{\Letter}  \and Shayok Chakraborty }

\begin{document}

\tocauthor{Chutian Jiang, Hansong Zhou,Xiaonan Zhang,  Shayok Chakraborty}
\toctitle{FedAR: Addressing Client Unavailability in Federated Learning}

\maketitle

\begin{center}
	{Department of Computer Science, Florida State University, Tallahassee, FL, USA}\\
	\texttt{cj20cn@fsu.edu, hz21e@fsu.edu, xzhang@cs.fsu.edu, shayok@cs.fsu.edu}
\end{center}

\begin{abstract}

	Federated learning (FL) enables clients to collaboratively train machine learning models under the coordination of a server in a privacy-preserving manner.  One of the main challenges in FL is that the server may not receive local updates from each client in each round due to client resource limitations and intermittent network connectivity.  The existence of unavailable clients severely deteriorates the overall FL performance.  In this paper, we propose \textit{FedAR}, a novel client update \textit{A}pproximation and \textit{R}ectification  algorithm for FL to address the client unavailability issue. FedAR can get all clients involved in the global model update to achieve a high-quality global model on the server, which also furnishes accurate predictions for each client. To this end, the server uses the latest update from each client as a surrogate for its current update. It then assigns a different weight to each client's surrogate update to derive the global model, in order to  guarantee contributions from both available and unavailable clients. Our theoretical analysis proves that FedAR achieves optimal convergence rates on non-IID datasets for both convex and non-convex smooth loss functions. Extensive empirical studies show that FedAR comprehensively outperforms state-of-the-art FL baselines including FedAvg, MIFA, FedVARP and Scaffold in terms of the training loss, test accuracy, and bias mitigation. Moreover, FedAR  also depicts impressive performance in the presence of a large number of clients with severe client unavailability.


\end{abstract}

\section{Introduction}

Federated learning (FL) allows multiple clients to collaboratively learn a powerful global {machine learning} model without {sharing} the training data with the server. As a privacy-preserving and communication-efficient distributed learning framework, FL has garnered substantial research {attention} and has surged as a key enabler of distributed intelligence in many real-world applications, such as next-word prediction on mobile keyboards \cite{hard2018federated} and medical record analysis in digital health \cite{brisimi2018federated}. In the vanilla FL algorithm, known as FedAvg \cite{mcmahan2017communication}, the server distributes the current global model to {all the} clients in each round, which serves as the basis for running several steps of {stochastic gradient descent (SGD)} on the local data for each client. The local updates are then sent back to the server to update the global model. This process is iterated until the global model {converges}.

In  FL, clients can be diverse, ranging from medical wearables and IoT devices to smartphones. Many of these clients operate as low-power devices and communicate over wireless networks. This presents a challenge to FedAvg, as clients may abort training midway due to issues like low battery levels or incoming calls \cite{bonawitz2019towards,gu2021fast,kairouz2021advances,mcmahan2017communication}. As a result, clients may fail to return their trained local updates to the server, especially when the communication from the clients to the server is hampered by poor channel quality and intermittent connectivity (also referred to as \textit{unavailable / non-participating clients} or the \textit{partial client participation problem}). In FedAvg, the inability to receive local updates from unavailable clients can cause a serious delay and it can even discard these updates when deriving the global model to maintain learning efficiency \cite{shu2021flas,wang2023fedgs,yan2024federated,zhu2021delayed}. Missing the expected local updates introduces an undesired bias against unavailable clients \cite{abdelmoniem2023refl,wang2023fedgs}. This will result in the global model overfitting the characteristics of consistently available clients, thereby diminishing its performance for clients that participate less frequently and reducing its overall generalization capability \cite{chen2022towards,horvath2021fjord,huang2022learn,mendieta2022local,zhou2022you}.

The primary goal of this paper is to develop and validate an efficient FL algorithm termed \textit{F}ederated Learning with local update \textit{A}pproximation and \textit{R}ectification (FedAR), which addresses the partial client participation problem. We first study the contributions of the latest observed local updates from unavailable clients to the global update. Our observation reveals that unavailable clients with varying inactive rounds exert diverse positive influences on the global update.  Motivated by this insight, we propose a novel server-side aggregation strategy that incorporates local updates from unavailable clients in the global update. More importantly, our framework does not require any additional computation at the clients or introduce any extra communication between the clients and the server.  FedAR utilizes the latest update from each client observed by the server as a surrogate of its current update, which is then used in updating the global model.  Moreover, we devise an innovative  weighting scheme to  accommodate the variable influence on the global model from local updates of clients with differing inactive rounds.  {We} slightly magnify the contributions from unavailable clients {(based on the number of inactive rounds) in addition to the contributions from the available clients, to update the global model}.  To achieve this, we design the weight as a mildly increasing function of the {number of} inactive rounds {of each} client.  This {strategy} enables the server to include {the local} data distribution information from  unavailable clients {in updating the global model, thereby} circumventing the bias against {these} clients.  Lastly, unlike traditional FL, FedAR does not {assume} that the server is aware of the total number of clients {in advance}.  Instead, it dynamically counts the number of clients who get involved in the global model update, which better reflects real-world application scenarios.  In light of the above discussion, we summarize our key contributions {in} this paper as follows:
\begin{itemize}
	\item We propose FedAR, a novel FL algorithm that {addresses} the client unavailability issue. FedAR unevenly weighs the contributions from both available and unavailable clients in the global model update based on the number of their inactive rounds. Moreover, FedAR does not necessitate any additional computation at the clients, nor does it demand any extra communication between the clients and the server. It  does not require all clients to participate in FL in the first round either.
	\item We theoretically provide a convergence guarantee for FedAR {for} both convex and non-convex {smooth} loss functions on non-IID datasets across clients.
	\item We evaluate the performance of FedAR {on} three real-world datasets MNIST, CIFAR-10, and SVHN. Compared {to} the vanilla and the state-of-the-art {FL baselines}, FedAvg, MIFA, FedVARP, and Scaffold, FedAR can achieve a $75\%$ improvement in test accuracy and a $50\%$ reduction in training loss {in the best case}. Moreover, we empirically show that FedAR can better mitigate the bias against unavailable clients, as evidenced by the observation that the derived global model generates more accurate predictions for clients who have been intermittently inactive during the training process. FedAR {also demonstrates impressive performance in the} presence of a large number of clients with {severe client unavailability}.
\end{itemize}

\section{Related Work}
 {One} of the main challenges {of the vanilla FL algorithm, FedAvg,} {is} the intermittent unavailability of clients. Specifically, the server will not update the global model until receiving local updates from all clients, which results in considerable training delay in the presence of client unavailability. Client sampling can be used as a remedy to this issue, where some clients are selected to participate in the global model update. The common client sample strategies include random sampling, significant sampling, and cluster sampling. Random sampling \cite{mcmahan2017communication} selects clients at random whereas importance sampling \cite{cho2020bandit,cho2022towards,luo2022tackling} selects the most valuable clients in terms of data quantity, communication time, and local training results. In cluster sampling \cite{briggs2020federated,fraboni2021clustered,ghosh2020efficient},  clients are first divided into groups based on sample size,  model similarity etc.; the clients in each group are then selected for global update.  All these sampling strategies engage only available clients but ignore unavailable clients in the global update. Consequently, the global model biases towards {the} available clients that are selected repetitively \cite{mohri2019agnostic}, which would undermine the FL performance.

A body of research addresses the client unavailability issue by incorporating stale updates from unavailable clients into the training process, {such as the Memory-augmented Impatient Federated Averaging (MIFA) algorithm \cite{gu2021fast} and the Federated VAriance Reduction for Partial Client Participation (FedVARP) algorithm \cite{jhunjhunwala2022fedvarp}. Their major differences with FedAR are listed in Table. \ref{difference}.  In particular, seeking to maximize non-IID data coverage, MIFA gives equal weightage to updates from both available and unavailable clients,  making it a biased scheme. Even worse, MIFA requires all clients to participate in the first {training} round, which is an unrealistic assumption. FedVARP allocates higher weights to the updates from available clients than to the updates from unavailable clients. It also attempts to reduce the variance to available clients caused by the partial client partition, which, however, is not empirically demonstrated. Similar to both MIFA and FedVARP, the FedAR algorithm reuses the latest observed update for each client as an approximation of its current update. Different from MIFA, FedAR formulates a novel weighting scheme to efficiently involve unavailable clients with various inactive rounds in the global model update. Moreover, FedAR does not require all clients to participate in FL in the first training round. Motivated by \cite{wang2022unified},  FedAR assigns higher weights to the updates of the unavailable clients with a larger number of inactive rounds, i.e., we amplify the local updates from unavailable clients, which is contrary to FedVARP. {Our experimental results show the {efficacy of FedAR in terms of overall convergence, test accuracy and bias mitigation, compared to relevant baselines}.

\begin{table}[!hp]
	\centering
	\begin{tabular}{|llll|}
		\hline
		\multicolumn{1}{|l|}{}                                       & \multicolumn{1}{l|}{MIFA}                          & \multicolumn{1}{l|}{FedVARP}                              & \multicolumn{1}{l|}{FedAR}                             \\ \hline
		\multicolumn{4}{|c|}{Enhance the FL efficiency with uncertain availability of clients}                                                                                                                                                 \\ \hline
		\multicolumn{1}{|l|}{\scriptsize \begin{tabular}[l]{@{}l@{}}
				                                 Issue \\ addressed
			                                 \end{tabular} } & \multicolumn{1}{l|}{{\begin{tabular}[l]{@{}l@{}}
						                                                                        maximize \\ data coverage\end{tabular}}}   & \multicolumn{1}{l|}{{\begin{tabular}[l]{@{}l@{}}
						                                                                                                                                          reduce  variance of \\ available local updates
					                                                                                                                                          \end{tabular}}}          & \begin{tabular}[l]{@{}l@{}}
			                                                                                                                                                                     {reduce bias against} \\ {unavailable local updates}\end{tabular} \\ \hline
		\multicolumn{1}{|l|}{{\begin{tabular}[l]{@{}l@{}}
						                      Rationale on \\local updates\\
					                      \end{tabular}}}            & \multicolumn{1}{l|}{{\begin{tabular}[l]{@{}l@{}}
						                                                                        all have the \\ same  contribution \\
					                                                                        \end{tabular}}}   & \multicolumn{1}{l|}{{\begin{tabular}[l]{@{}l@{}}
						                                                                                                                 available ones  have \\ higher  contributions \\
					                                                                                                                 \end{tabular}}}          & \begin{tabular}[l]{@{}l@{}}
			                                                                                                                                            unavailable ones can also \\ have contributions\\
		                                                                                                                                            \end{tabular}                                             \\ \hline
		\multicolumn{1}{|l|}{{\scriptsize Solution}}                 & \multicolumn{1}{l|}{{\begin{tabular}[l]{@{}l@{}}
						                                                                                    allocate the same  weight \\ to  all local updates
					                                                                                    \end{tabular}}} & \multicolumn{1}{l|}{{\begin{tabular}[l]{@{}l@{}}
						                                                                                                                           allocate higher  weights \\ to available local updates \\
					                                                                                                                           \end{tabular}}} & \begin{tabular}[l]{@{}l@{}}
			                                                                                                                                             allocate higher weights  to unavailable \\ local updates  with higher contributions
		                                                                                                                                             \end{tabular}        \\ \hline
		\multicolumn{1}{|l|}{\scriptsize \begin{tabular}[l]{@{}l@{}}
				                                 All clients \\ assumption
			                                 \end{tabular} } & \multicolumn{1}{l|}{\begin{tabular}[l]{@{}l@{}}
				                                                                       must respond  in \\ the first round
			                                                                       \end{tabular}}    & \multicolumn{2}{l|}{\begin{tabular}[l]{@{}l@{}}
				                                                                                                               \textbf{not necessarily} respond in the first round
			                                                                                                               \end{tabular}}                                                                      \\ \hline
	\end{tabular}
	\caption{Comparison of FedAR with MIFA and FedVARP}
	\label{difference}
\end{table}

\section{Problem Setup}
We consider that a set of clients $\mathcal{N}=\{1,2, \cdots, N\}$ with restricted power and computational resources collaborate with a server to execute FL over $T$ rounds. The datasets for local training are subject to non-IID distributions. The clients and the server iteratively communicate over wireless networks to obtain a global model $w$ aiming at minimizing the global loss function:
\begin{equation}
	\min f(w)=\frac{1}{{N}}\sum\nolimits^{N}_{i=1}f_i(w),
\end{equation}
where $f_i(w)$ is the loss function for client $i$.

\subsection{Basic Algorithm of FL}
{We begin by recalling the vanilla FL setting in FedAvg. } In round $t-1, t \in \{1, \cdots, T\}$,  the server broadcasts the global model ${w}_{t-1}$ to {all the} clients. Each client $i \in \mathcal{N}$ uses its own private dataset to execute $K$ {steps of} Stochastic Gradient Descent (SGD) for the local update. For each step $k \in {K}$:
\begin{equation}\label{eq2}
	w^i_{t,k+1} = w^i_{t-1,k}-\eta_{t-1} \nabla f_i(w^i_{t-1,k}),
\end{equation}
where $\eta$ is the local learning rate and  $\nabla f_i(\cdot)$ represents the  gradient. Each client then sends back its local update to the server; the server aggregates all {the client updates to derive the global model as:}
\begin{equation}
	w_t=\frac{1}{{N}}\sum\nolimits^{N}_{i=1} w^i_{t, K}.
\end{equation}
\subsubsection{Problem in FedAvg.}
Practically, due to the limited resources of each client and the intermittent network connectivity, the server may not receive the local updates ${w}^i_{t, K}$ from all the clients; {these clients are called \textit{unavailable} / {\textit{non-participating}} clients}. {Due to this}, FedAvg delays or even aborts the local updates from unavailable clients during the global update, causing an undesirable bias against these unavailable clients. However, the local updates from the unavailable clients also contain valuable information, which can be useful in global model updates.  We conduct {a toy} experiment {on a simple, restricted setup} to demonstrate this idea and provide motivation for our approach.
\begin{figure}[H]
	\centering
	\includegraphics[width=0.8\linewidth]{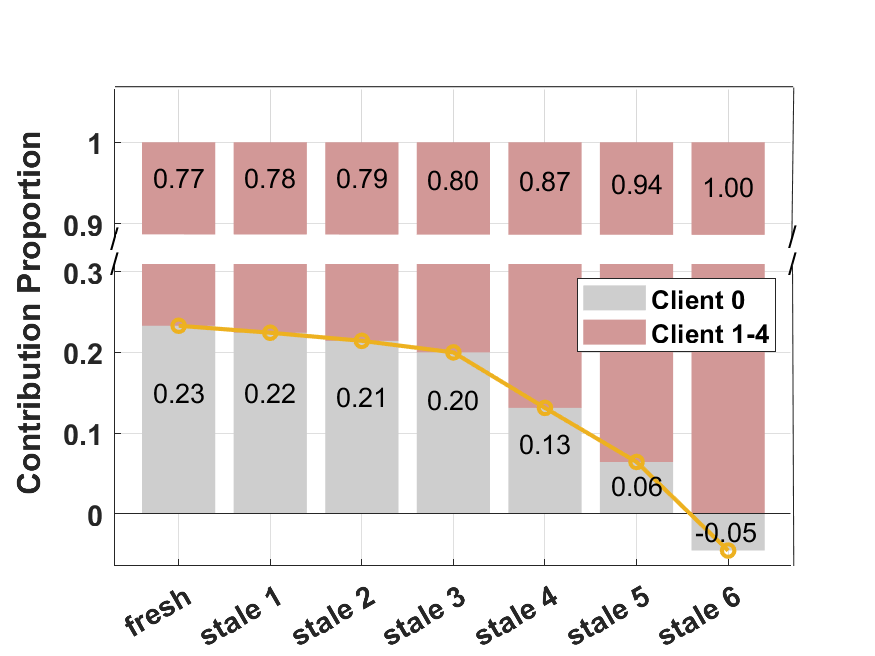}
	\caption{Contribution of each client {to the global model}. {``stale $i$'' denotes that Client $0$ has been inactive for the last $i$ rounds. ``fresh'' denotes that all the clients are active for all the $9$ rounds. A high staleness level indicates more inactive rounds}}\label{fig:Motivation3}
\end{figure}
\subsection{Motivation}

Let us assume a standard FL setting where  $5$ clients (numbered $0$ through $4$) collaborate with a central server on a classification task using the CIFAR-10 dataset \cite{CIFAR10}. The server and clients execute a total of $9$ rounds of communication. We conduct $7$ different experiments, as shown  by the vertical bars in Fig.  \ref{fig:Motivation3}. In all the experiments, client $1$ to client $4$ are always available across all the $9$ rounds of communication. Client $0$, conversely, becomes inactive after a certain number of rounds in each experiment. In Fig. \ref{fig:Motivation3}, the term ``stale $i$” refers to client $0$ being active for the initial $9-i$ rounds and then inactive for the subsequent $i$ rounds. For instance, ``stale $3$” indicates that client $0$ is active from rounds $1$ to $6$ but inactive during rounds $7$ to $9$. In this case, we aggregate the most recent local updates (from the $9^{th}$ round) for clients $1$ to $4$ and the local update from the $6^{th}$ round for client $0$ (last active round) to update the global model.  ``fresh'' denotes the case where all the $5$ clients were available across all the $9$ rounds of communication. After the $9^{th}$ round, we use the Shapley Value (SV) \cite{shapley1953value} to quantify the contribution of each local update to the global model. Shapley value is a classical concept in cooperative game theory, and it is extensively used to evaluate client contributions in FL \cite{soltani2023survey,song2019profit,wang2019measure}. We {compute} each client's SV based on the global model's test accuracy, which {is obtained} by different combinations of the local client updates for the different experiments. We sum up all the SV and represent the contribution of client $0$ and clients $1$ to $4$ {as a percentage}; the larger the value, the greater the contribution. {From} Fig. \ref{fig:Motivation3}, it can be observed that as the staleness level of client $0$ increases (larger number of inactive rounds), {its contribution to the global model} {(height of the gray bar)} decreases. {At} stale $6$, the contribution of client $0$ is negative, meaning that its local update has an adverse effect on the global model.
Based on the above toy experiment, we draw the following conclusions:

\begin{itemize}
	\item The stale local updates from unavailable clients can still contribute to the global model (as evident from the gray bars in stale $1$ through stale $5$).
	\item The contribution of the stale local updates decreases with increasing staleness level, suggesting it may be necessary to assign higher weights (during the global update) to stale local updates with more inactive rounds.
	\item An excessively high staleness level is detrimental to the performance of the global model. In the global update, it may not be necessary to include these local updates.
\end{itemize}

Given these observations, we propose our FedAR algorithm, as detailed below.

	{\section{{FedAR Algorithm}}}
FedAR is designed as a simple and effective algorithm by involving local updates of unavailable clients in the global model update on the server. In addition, given that a client's unavailability leads to decreased contributions, we assign weights to different local updates accordingly. Our goal is to enhance FL performance by efficiently involving updates from all clients in the global model update. Specifically, FedAR consists of two components: \textit{local update approximation} and \textit{local update rectification}. In each round, the server sets a maximum waiting time for the local updates from all clients. When the maximum waiting time is reached, the server estimates the local updates that would be obtained from the unavailable clients.  The weighted average over all the local updates is then performed to derive the global model for the next round. We describe our system in detail next.

\subsubsection{Local Update Approximation}
To approximate the local updates, the server maintains an update-matrix ${\mathbf{G}[t]} = [{G}_1[t]; \cdots; {G}_i[t];  \cdots; {G}_{N}[t]]$ saving its most recent observed local updates from all clients. Initially, $\mathbf{G}[0]$ is a zero matrix.  In round $t$,  ${G}_i[t]$ will only be replaced if the server obtains the client $i$'s local update ${w}^i_{t, K}$. Otherwise, ${G}_i[t]$ will not change.  Let $\mathcal{A}(t) \subset \mathcal{N}$ represent the set of available clients whose updates are successfully received by the server in round $t$.   Mathematically, we have,
\begin{equation}\label{MIFABuffer}
	G_i[t] = \begin{cases}
		\frac{1}{\eta_t} (w_t-w^i_{t,K}) & \text{if client }i \in \mathcal{A}(t) \\

		G_i[t-1]                         & \text{otherwise.}
	\end{cases}
\end{equation}

FedAR {uses} ${G}_i[t] \in \textbf{{G}}[t]$ as the estimates for local updates while deriving the global model. The global model is thus able to include the data distribution from the unavailable clients, which will help mitigate the bias against them.

\subsubsection{Local Update Rectification}
Fig. \ref{fig:Motivation3} shows that stale local updates with different inactive rounds have various contributions to the global model, inspiring us to weigh local updates during the global update. We propose to assign weights to local updates based on the number of their inactive rounds, and by doing so, we expect to enhance the contributions from unavailable clients and further mitigate the bias.

	{Formally}, the server maintains an update-array ${\tau}(t-1) = [\tau(1,t-1), \cdots, \tau(i,t-1), \cdots, \tau(N,t-1)]$ to record the number of inactive rounds for all clients. ${\tau}(0)$ is initialized as a zero array. In round $t$, if the update from client $i$ is received, the server resets $\tau(i,t)$  to $0$. Otherwise, the server increases $\tau(i,t-1)$ by $1$ to get $\tau(i,t)$. We express $\tau(i,t)$ as:
\begin{equation}\label{tau}
	\tau(i,t) = \begin{cases}
		0               & \text{if client }i \in \mathcal{A}(t) \\
		\tau(i,t-1) + 1 & \text{otherwise}.
	\end{cases}
\end{equation}
Based on ${\tau}(i, t)$, we design a weight function $\psi_{i,t}$ to {quantify} the contribution from client $i$ to the global update. The general expression of $\psi_{i,t}$ is {given as:}
\begin{gather}\label{compensationscore}
	\psi_{i,t} = \begin{cases}
		0                              & \text{if  } \tau({i,t}) \geq g(t) \\
		\min([\tau(i,t) +1]^{\rho}, 2) & \text{otherwise}.
	\end{cases}
\end{gather}
If the client $i$ is available at round $t$, i.e.,  $ \tau({i,t}) = 0$, we have $\psi_{i,t}= 1$, which aligns with FedAvg. We introduce $g(t)$ to prevent local updates with many inactive rounds from negatively impacting the global model and to remove such updates from the current global update.  $g(t)$, as a function of round $t$, is different {based on whether we are optimizing a convex loss function or a non-convex loss function}. We will discuss it in more detail in Section $5$ (Theoretical Analysis).

	{Since unavailable clients with more inactive rounds contribute less to the global update, we assign them higher weights to increase their contributions, as shown in Eq. (\ref{compensationscore}).  However, an extremely high weight $\psi_{i,t}$ will cause the unavailable clients to dominate the global model update, which would induce bias against available clients. We therefore introduce the hyperparameter $\rho \in [0,1]$ in Eq. (\ref{compensationscore}) to restrict the growth of $\psi_{i,t}$. We also set the maximum value  of $\psi_{i,t}$ ($\psi_{max}$) to $2$} to guarantee the convergence of FedAR. Please refer to the Appendix for more details on the convergence analysis}.

\subsubsection{Global Model Update}
{Clients arbitrarily participate in global model update in each round due to their limited resources and intermittent network connectivity. Hence, the server does not know the exact number of clients in advance; instead, it dynamically counts the clients that contribute to the global model update, i.e. those clients that are either available or unavailable but not too stale.
	Suppose there are $N_t$ contributing clients in round $t$.  With  $G_{i}[t]$ and $\psi_{i,t}$, FedAR updates the global model as follows:}

\begin{equation}\label{eq2_update}
	w_{t+1} = w_t-\frac{\eta_t}{N_t}\sum\nolimits_{i=1}^NG_i[t]\psi_{i,t}.
\end{equation}

{Combined with Eq. (\ref{MIFABuffer}), Eq. (\ref{eq2_update})  ensures that the update matrix $G_i[t]$ always reflects the most recent client updates, while being able to reasonably consider the contributions of all clients when the global model is updated.} In addition, although Eq. (\ref{eq2_update}) seems to have all clients in the global model update, some clients do not get involved. They are either the clients that have never participated in FL, i.e., $G_i[t] = 0$,  or the clients that have been inactive for many rounds, i.e., $\psi_{i,t} = 0$.  Hence, Eq. ({\ref{eq2_update}}) aligns with our idea of engaging only the contributing clients in the global model update.

Algorithm 1 shows the details of FedAR. We use the ``temporary client set \(\mathcal{E}\)'' to include the clients that have ever participated in the global model update in Line 3. Initially, $N_t$ is the number of clients in $\mathcal{E}$. When the client $i$ has been inactive for many rounds, i.e., \(\psi_{i,t} = 0\), it will be excluded from the global model update, i.e., $N_t =N_t-1$ in Line 11. Ultimately, $N_t$ counts the contributing clients as in Eq. (\ref{eq2_update}).

\begin{algorithm}[tb]
	\small
	\caption{FedAR}
	\label{alg:algorithm}
	\textbf{Input:} initial $w_0$, learning rate $\eta_t $,  local step $K$,  total round number $T$, total client number $N$

	\textbf{Output}: The derived global model $w_T$

	\textbf{Server executes:}
	\begin{algorithmic}[1] 
		\STATE Initialize $\psi_{i,1}=1$, $\tau(i,1)=0$,  and $G_i[0]=0$, $\forall i$, temporary client set  $\mathcal{E}$

		\FOR{$t$=1,2, $\cdots , T$}
		\STATE  $\mathcal{E}$ $\leftarrow$ $\mathcal{E}$ $\cup$ \{\text{{new active client}}\}, $N_t= \vert $$\mathcal{E}$$ \vert$.
		\FOR{$i$=1,2 $\cdots , N$ \textbf{in parallel}}
		\IF{client $i$ is available}
		\STATE $G_i[t]$ $\gets$ DeviceUpdate($i$,$w_t$)
		\STATE $\tau(i,t)=0$
		\ELSE
		\STATE $\tau(i,t)=\tau(i,t)+1$
		\ENDIF
		Calculate the $\psi_{i,t}$ by Eq. (\ref{compensationscore})
		\IF{$\psi_{i,t}=0$}
		\STATE $N_t= N_t -1$
		\ENDIF
		\ENDFOR
		\STATE $w_{t+1}$ $\gets$ $w_t-\frac{\eta_t}{N_t}\sum^{{N}}_{i=1}G_i[t]\psi_{i,t}$
		\ENDFOR

	\end{algorithmic}
	\textbf{DeviceUpdate($i$,$w_t$)}:
	\begin{algorithmic}[1]
		\STATE $w_{t,0}^i \leftarrow w_t$
		\FOR{k = 0,1,$\cdots, K-1 $}
		\STATE $w^i_{t,k+1}\leftarrow w^i_{t,k}-\eta_t{\nabla} f_i(w^i_{t,k})$
		\ENDFOR
		\STATE Return $\frac{1}{\eta_t}(w_t-w^i_{t,K}) $
	\end{algorithmic}
\end{algorithm}

{Regarding privacy enhancements in FL, FedAvg  suggests that Differential Privacy (DP) can improve data privacy performance. } However, our work is not primarily focused on privacy protection, and as such, an in-depth examination of this topic will not be included in our current research.

\section{Theoretical Analysis of FedAR}
 {In this section, we analyze the convergence of the proposed FedAR for convex and non-convex smooth loss functions.}

\subsection{Convex Loss Function}
To analyze the convergence of  FedAR for a convex loss function, we make the following assumptions regarding $f_i(w), i = 1, 2, \cdots,  {N}$.

\noindent\textbf{Assumption 1: L-smoothness.} The loss function $f_i(w)$ is L-smooth. That is:  for all $x, y$ $\in$ $\mathbb{R}, f(x)-f(y)\leq \langle \nabla f(y),x-y\rangle +\frac{L}{2}\| y-x\|^2$ with  $L >0$.\\
\textbf{Assumption 2: $\mu$-strong convex.} The loss function $f_i(w)$ is $\mu$-strong convex. That is:  for all $x, y$ $\in$ $\mathbb{R}, f(x)-f(y)\geq \langle \nabla f(y),x-y\rangle +\frac{\mu}{2}\| y-x\|^2$ with  $\mu >0$.\\
\textbf{Assumption 3: Variance bound.} The variance of the unbiased estimator of $\nabla f_i(w)$ in round $t$ is upper bounded, where $\mathbb{E}\{\| \tilde{\nabla}f_i(w) -{\nabla}f_i(w) \psi_{i,t}\|^2\} \le \sigma^2$.\\

\noindent\textbf{Theorem 1:} Suppose the objective loss function $f_i(w)$ satisfies Assumptions 1 to 3, $\tau_{max} \leq g(t)$. By setting the learning rate $\eta_t=\frac{4}{\mu (t+a)}$ and constant $a =100(\frac{L}{\mu})^{1.5}$, after $T$ rounds, FedAR satisfies:
\begin{align*}
	\mathbb{E}[f( & \overline{w}_T)]-f(w_*)\mathcal=\mathcal{O}(\frac{\sigma^2(1+\overline{\tau}_T)}{\mu K{N}T}) \\
	              & +\mathcal{O}(\frac{F+\|w_1-w_*\|^2+\tau^2_{max}L\sigma^2{N}\psi_{max}}{K\mu^3T^2}),
\end{align*}
where $\tau_{max}$ is the  maximum number of $\tau(i,t)$ over all clients and rounds. $g(t)= t_0 +\frac{1}{b}t$ for a constant $t_0 >0$ and  $ b >2 $,
$F= LK{N}D+L(K-1)^2\cdot (D{N}^2+\frac{\sigma^2}{K})$,
$\overline{w}_T = \frac{\sum_{t=1}^T(t+a-1)(t+a-2)w_t}{W_T}$, $W_T = \sum_{t=1}^T(t+a-1)(t+a-2)$,
$\overline{\tau}_T = \frac{1}{{N}(T-1)}\sum^{T-1}_{t=1}\sum^{N}_{i=1} \tau(i,t)$, and $D= \frac{1}{{N}}\sum^{N}_{i=1}\|\nabla f_i(w_*)\|^2$.

\noindent\textbf{Remark 1.}
In Theorem 1, both the first and the second terms tend to zero as $T$ increases, indicating that FedAR converges at the rate of $\mathcal{O}({{1}/{T}})$. The first term's convergence is related to the average inactive round number $\overline{\tau}_T$. We can find that too high a value of $\overline{\tau}_T$ will negatively impact convergence, which is consistent with our observation in Section 3.2 (Motivation). Also, convergence is adversely affected when most clients remain unavailable for a long time, i.e., a large $\overline{\tau}_T$. Besides, we observe that weight function $\psi$ has a relatively negligible effect on the convergence rate. This can be attributed to our restriction on $\psi$ in Eq. (\ref{compensationscore})  to prevent it from becoming excessively large with an increase of $\tau$. This is because a larger $\psi$ could lead to the dominance of clients with more inactive rounds during the global model update.

\subsection{Non-Convex Loss Function}
To analyze the convergence of  FedAR for a non-convex smooth loss function, we make the following assumptions regarding $f_i(w), i = 1, 2, \cdots, {N}$.

\noindent\textbf{Assumption 4: Hessian Lipschitz}. The Hessian of a twice differentiable function $f$: $\mathbb{R}^d \rightarrow \mathbb{R}$ is $\lambda $-Lipschitz continuous if $\left \| \nabla^2 f(x)-\nabla^2 f(x)  \right\|\leq  \lambda \left \| x-y \right\|$ for all $x, y$.\\
\textbf{Assumption 5: Gradient noise}.  The noise of the local stochastic gradients in round $t$ is upper bounded by a constant $\delta$: $\left \| \tilde{\nabla} f_i(w)-\nabla f_i(w) \right\| \leq \delta$.\\
\textbf{Assumption 6: Gradient dissimilarity}.  $\exists$ $\alpha >0$   and $\beta_i >0 $: $\left \| \nabla f_i(w) \right\|^2 \leq \alpha \left \| \nabla f_i(w) \right\|^2 +\beta_i >0$ and we define $\beta= \frac{1}{{N}}\sum^{N}_{i=1}\beta_i$.\\
\textbf{Assumption 7:} There exists a constant $v_i$ such that $\tau(i,t) \leq v_i$ for $\forall i \in \mathcal{N}$, and define $\overline{v}$ =$\frac{1}{N}\sum^N_{i=1}v_i$, $v_{max}=max_{i \in \mathcal{N}}v_i$.

\noindent\textbf{Theorem 2:} Suppose Assumptions 1 to 7 hold, set learning rate $\eta = \sqrt{\frac{{N}}{KTL(1+\overline{v})}}$, $T$ $\geq$ $max\{32\alpha L{N}K, 16L{N^5}K, \frac{8K{N}v^2_{max}(L^2+\lambda\delta {N^2})}{L}\}$, and $\tau_{max} \leq g(t) $. After $T$ rounds, FedAR satisfies:
\begin{align*}
	\mathbb{E} & [\|\nabla f(w_T)\|^2] \leq \mathcal{O}  ( R\sqrt{\frac{L(1+\overline{v})}{TK{N}}} (f(w_1)-f^*+\sigma^2)                       \\
	           & + \frac{\alpha \sigma^2 \overline{v} LK{N}^2\psi_{max}}{T}+\frac{ \sigma^2 \lambda \delta {N} \psi_{max}}{LT}+\frac{F_1}{T}),
\end{align*}
where $g(t)=\frac{1}{4}\sqrt{\frac{L}{(L^2+\beta\lambda {N})K{N}}}\times max\{\sqrt{t},\sqrt{t_0}\}$ for a constant $t_{0}>0$,
$F_1=(\alpha +1)(LK{N}\sigma^2\overline{v}  +LK{N}\sigma v_{max} \sqrt{\beta+\frac{\sigma^2}{K{N}}}) +\frac{(L^2+\lambda\delta {N}^2)\sigma v_{max}}{L}+(K-1)(2\beta+\frac{\sigma^2}{K})$, and $R = \frac{8\psi_{max}^2}{4\psi_{max}^2-1}$.

\noindent\textbf{Remark 2.}
In Theorem  2,  the convergence of FedAR for a non-convex smooth loss function is dominated by the first term, which converges at the rate of $\mathcal{O}(\sqrt{{1}/{T}})$. This dominant term is mainly influenced by the initial error$f(w_1)-f^*$, the variance bound $\sigma$, and the average upper bound of inactive round number across clients $\overline{v}$. In addition, we observe that weight function $\psi$ appears in the dominant term through the parameter $R$. Regardless of how $\psi$ changes, the value of $R$ tends towards a constant, and thus the impact produced by $\psi$ is not significant. Compared to $\psi$, $\tau$ and $N$ have a greater influence on the convergence via impacting the dominant term. We can draw similar conclusions as Theorem  1: as more clients continue to join the FL, more rounds are required to achieve convergence.  Meanwhile, the fact that $\tau_{max}$ is a major variable affecting convergence aligns with our initial observations in Section 3.2 (Motivation); that is,  the local updates with more inactive rounds negatively impact the global model's performance and further prevent the global model from converging. Thus, there must exist a critical value $g(t)$ as we express in Eq. (\ref{compensationscore}) to exclude those clients from the global update to ensure the model convergence, i.e.,. the clients whose inactive round number exceeds $g(t)$ will not be considered.

Please refer to our  Appendix for the proof of  Theorem 1 and  Theorem 2, as well as  Remark 3 on Theorem 2.

\section{Experiments and Evaluations}
In this section, we evaluate the {performance of} FedAR by conducting extensive experiments on a desktop with the GeForce RTX 3060 graphic card.

\subsection{Experimental Setup}
\noindent{\textbf{System Settings.}}
We conduct the FL experiments with one server and $100$ clients. Let $p_i$ denote the probability that client $i$ is available during any given round. The availability of all clients is independent,  with a minimum probability of $p_{min}$, indicating that the client availability probability varies from $p_{min}$ to $1$. This is a practical setting given that clients have their unique resource constraints and face distinct wireless environments. We examine both the challenging and mild client unavailability, where $p_{min} = 0.1$ and $0.5$, respectively. In summary, most clients are inactive for $5$ - $20$ rounds, with a few clients being inactive for more than $40$ rounds.

\noindent{\textbf{Data and Model.}}
We evaluate FedAR on three real-world datasets:  MNIST \cite{mnist},  CIFAR-10 \cite{CIFAR10}, {and} SVHN \cite{37648}. To ensure non-IID data distribution among all clients, we assume all datasets to be evenly distributed on all clients, and each client to contain only two classes of data. We use the logistic regression for MNIST, Lenet-5 for CIFAR-10, and Resnet-18 for SVHN as the local models. We set all experiments' initial local learning rate as $\eta_0=0.1$, local training step as $K=5$, local batch size as $64$, and hyperparameter as $\rho = 0.1$. We set weight decay as $0.001$ during the local SGD.

\noindent{\textbf{Baselines.}} We compare FedAR with {recent FL} baselines: (1) \textit{MIFA} \cite{gu2021fast}. It assigns the same weight to both available and unavailable clients; (2) \textit{FedVARP} \cite{jhunjhunwala2022fedvarp}. It assigns higher weights to available clients' updates, while the weights for unavailable clients remain unchanged; (3) \textit{FedAvg-IS}. It engages only available clients in global update using the FedAvg algorithm. The local updates are weighted by clients' availability probabilities; (4)  \textit{FedAvg (S=50)}. It involves at most half of available clients in the global update with the FedAvg algorithm.   Given $100$ clients, at most $50$ clients join the global update; and (5) \textit{Scaffold} \cite{karimireddy2020scaffold}. It  is a FL algorithm designed to improve the quality of global model updates by applying personalized control variate adjustments to each client; it does not consider client unavailability.

\subsection{Experimental Results\protect\footnotemark}
\footnotetext{For clear observation, we recommend viewing all figures about experimental results in color}
\noindent{\textbf{Overall Convergence Performance.}}
We evaluate the convergence performance of FedAR on different datasets in both the challenging and mild {settings} in Fig.  \ref{convergePerformance}. We find that FedAR has a similar convergence speed as FedAvg-IS, MIFA, and FedVARP. Notably, on CIFAR and SVHN datasets, the convergence speed of FedAR is markedly superior to that of Scaffold. This observation is consistent with our theoretical analysis that our designed weight function $\psi$ has negligible negative impacts on convergence.

When $p_{min}=0.1$, Fig. \ref{cifar_0.1} shows that  FedAR on CIFAR-10 reduces the training loss to 1.5 and attains the highest test accuracy of 44$\%$,  an enhancement of over 3$\%$ compared to baseline algorithms. Fig. \ref{SVHN_0.1} shows that FedAR is the only algorithm achieving a training loss below $1$ and a test accuracy over $70\%$ on SVHN.  When more clients are available, i.e., $p_{min}=0.5$, FedAR in Fig. \ref{cifar_0.5} greatly boosts the test accuracy to $46\%$ on CIFAR-10. Additionally, we find that FedAR consistently reaches a test accuracy of around $70\%$ in most training rounds on SVHN, and outperforms all the baselines.

\begin{figure}[t]
	\begin{subfigure}[b]{0.5\textwidth}\label{p1_results_mnist_logit_trainingloss}
		\centering
		\includegraphics[width=1\linewidth]{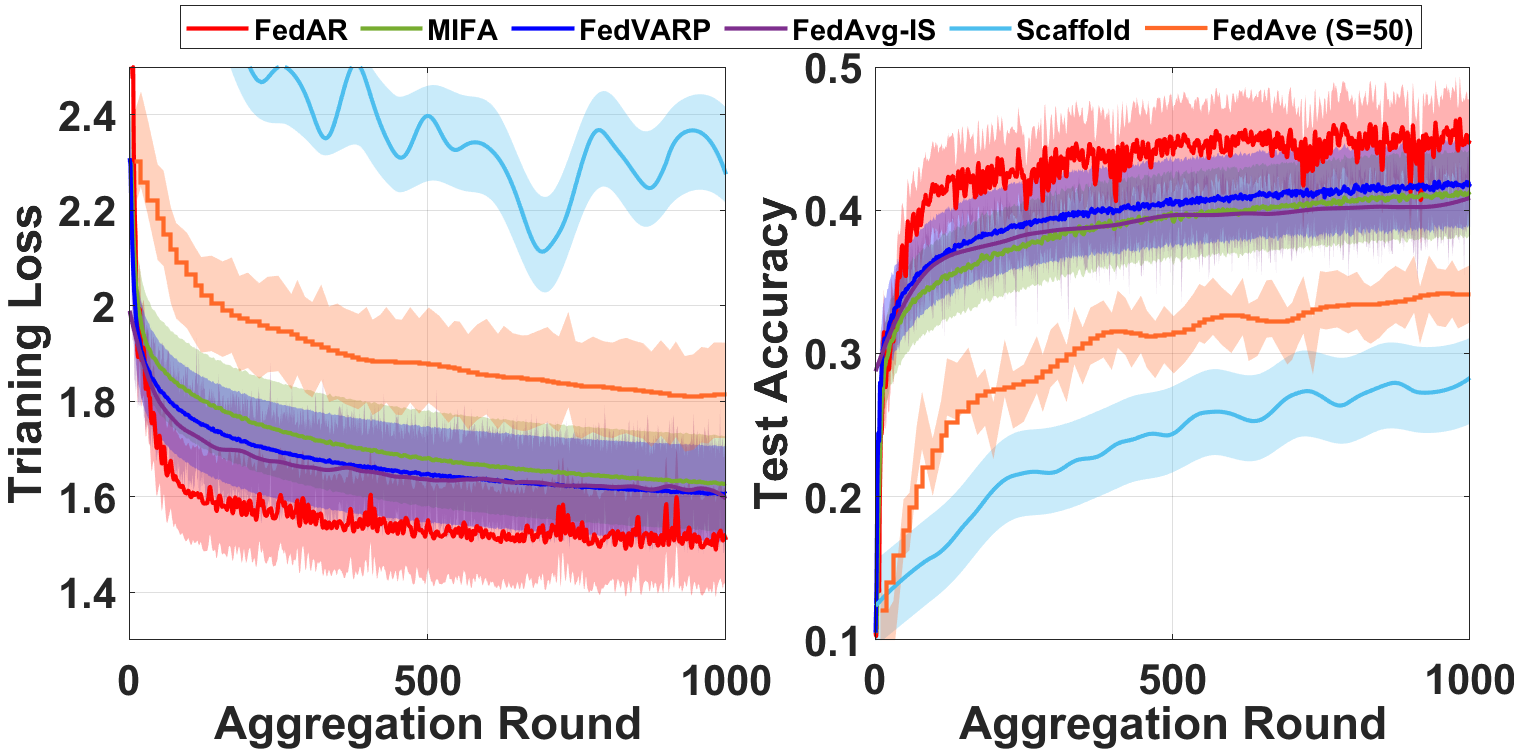}
		\caption{CIFAR-10: $p_{min} = 0.1$} \label{cifar_0.1}
	\end{subfigure}
	\begin{subfigure}[b]{0.5\textwidth}\label{p5_results_mnist_logit_trainingloss}
		\centering
		\includegraphics[width=1\linewidth]{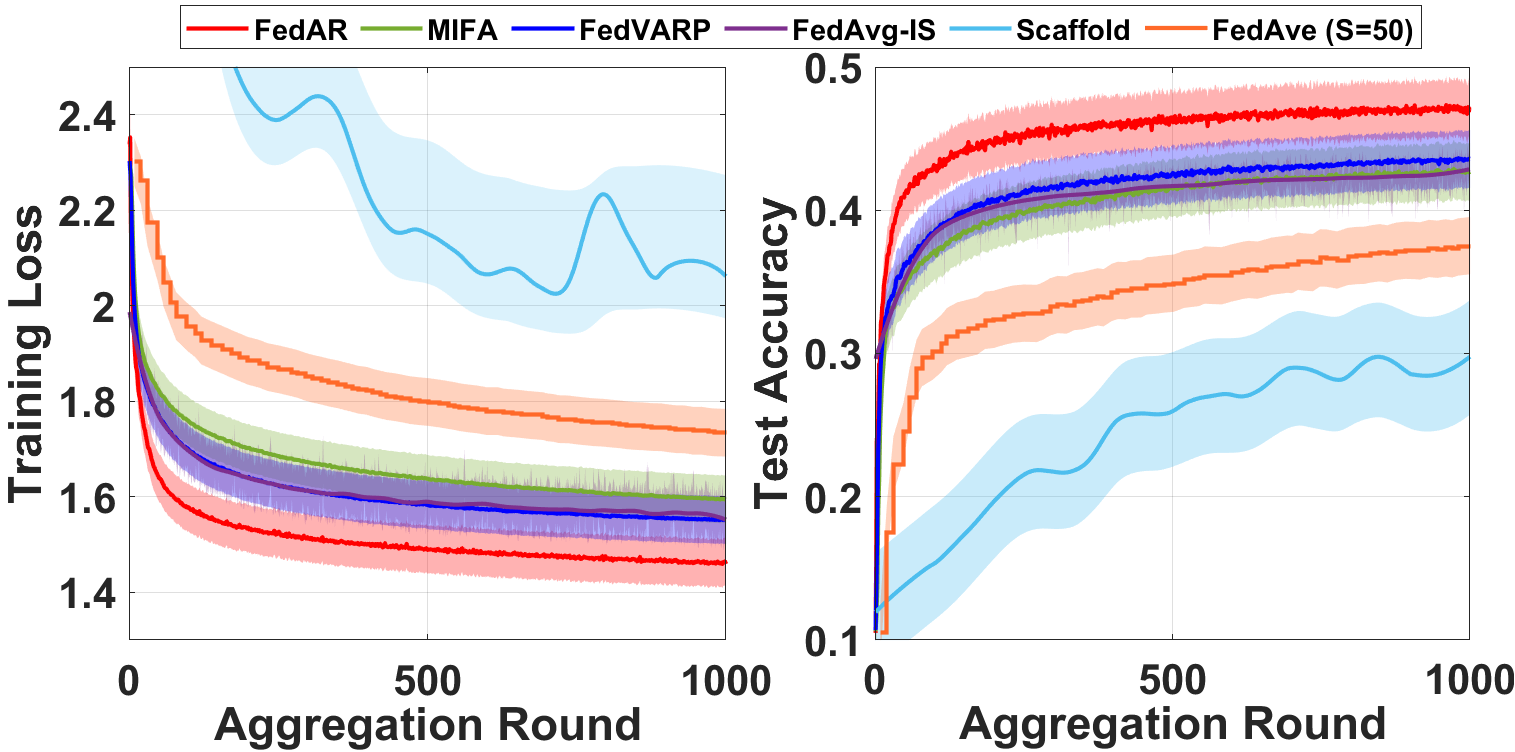} 
		\caption{CIFAR-10: $p_{min} = 0.5$} \label{cifar_0.5}
	\end{subfigure}%
	\\
	\begin{subfigure}[b]{0.5\textwidth}\label{p1_results_cnn_testacc}
		\centering
		\includegraphics[width=1\linewidth]{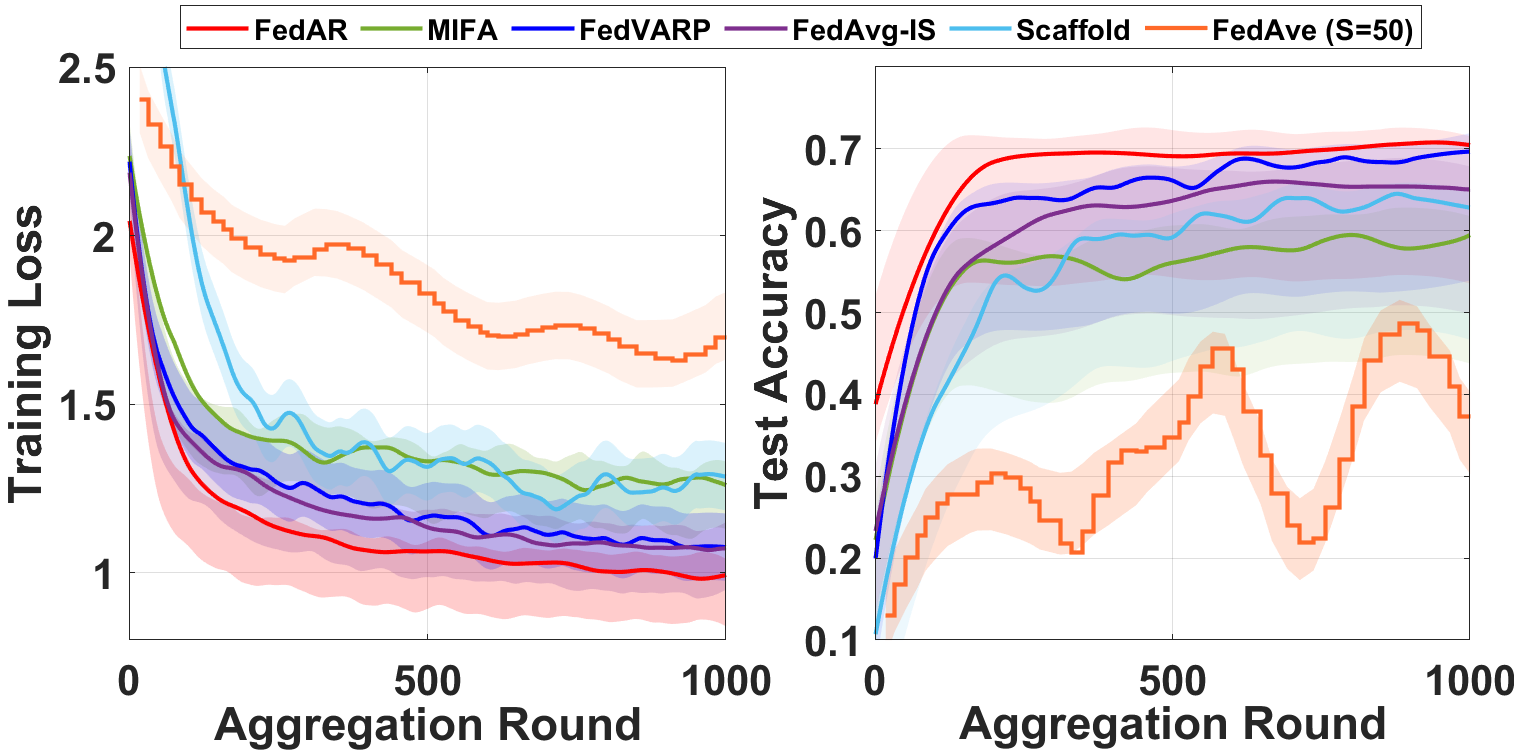}
		\caption{SVHN: $p_{min} = 0.1$}\label{SVHN_0.1}
	\end{subfigure}%
	\begin{subfigure}[b]{0.5\textwidth}\label{p5cnn_tc}
		\centering
		\includegraphics[width=1\linewidth]{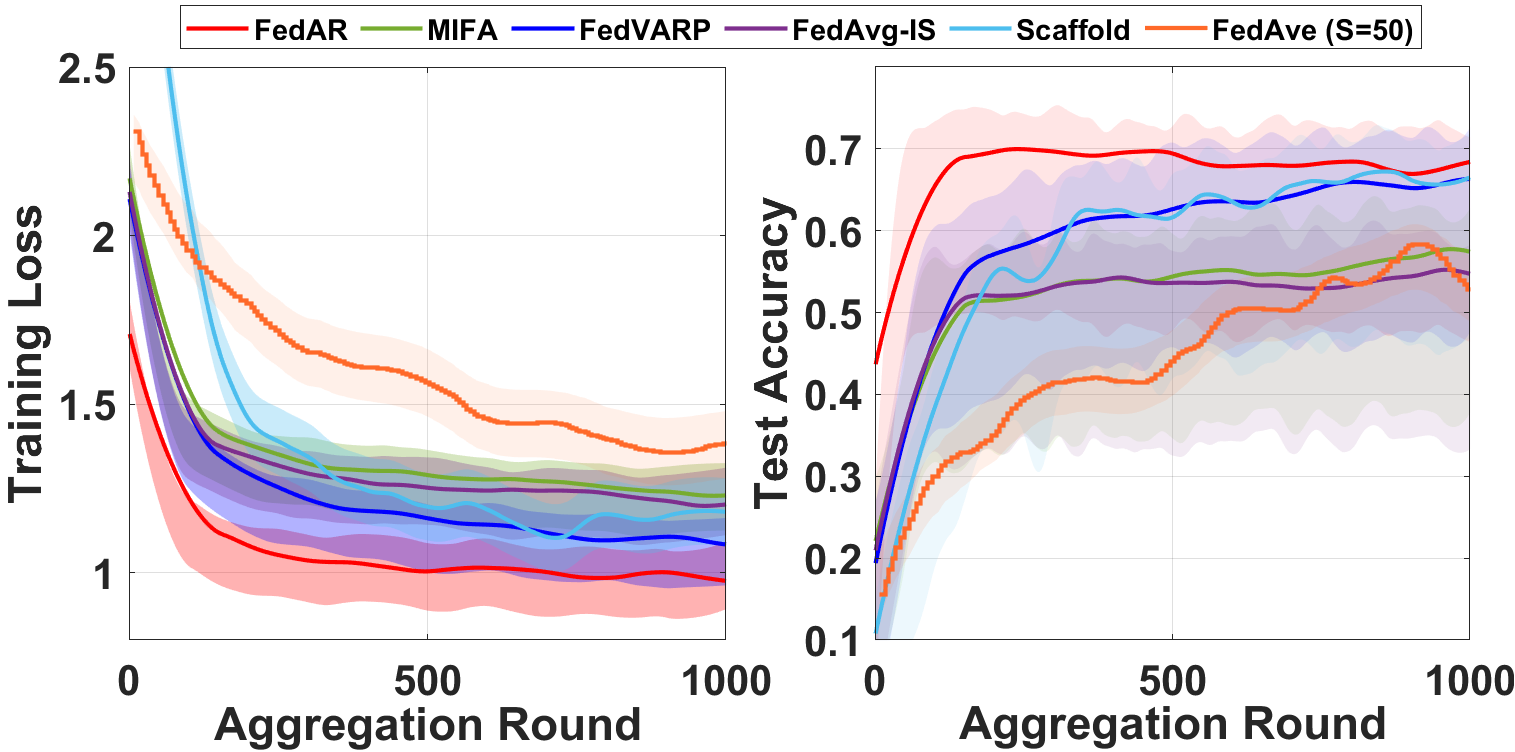}
		\caption{SVHN: $p_{min} = 0.5$} \label{SVHN_0.5}
	\end{subfigure}
	\caption{Convergence, training loss and test accuracy performance }\label{convergePerformance}
\end{figure}

\begin{table}[H]
	\renewcommand{\arraystretch}{1}
	\resizebox{1\columnwidth}{!}{
	\begin{tabular}{llllll}
		\hline
		\multicolumn{1}{c}{{Dataset}} & \multicolumn{5}{c}{Baselines}                                                                                   \\ \cline{2-6}
		                              & FedVARP                       & MIFA               & Scaffold          & FedAve(s=50)      & FedAve-IS          \\ \hline
		Cifar10;p=0.1                 & 1.15*10$^{-163}$              & 1.15*10$^{-194}$   & 1.87*10$^{-194}$  & 1.53*10$^{-110}$  & 4.36*10$^{-180}$   \\
		Cifar10;p=0.5                 & 0.0                           & 0.0                & 1.38*10$^{-246}$  & 3.46*10$^{-316}$  & 2.02*10$^{-285}$   \\
		SVHN;p=0.1                    & 0.00029                       & 1.49*10$^{-54} $   & 5.72*10$^{-35} $  & 1.63*10$^{-60} $  & 4.77*10$^{-45} $   \\
		{SVHN;p=0.5 }                 & {1.34*10$^{-52}$}             & {5.91*10$^{-111}$} & {1.71*10$^{-81}$} & {2.25*10$^{-77}$} & {1.96*10$^{-116}$} \\ \hline
	\end{tabular}
	}
	\captionof{table}{P-Value analysis of FedAR performance }\label{p_value}
\end{table}

We also conduct statistical tests of significance using paired t-test to assess whether the improvement in performance achieved by FedAR is statistically significant. We compare the test accuracy of FedAR against each of the baselines individually for both CIFAR-10 and SVHN, and for $p_{min}=0.1$ and $p_{min}=0.5$. The results are illustrated in Table \ref{p_value}; each entry
in the table denotes the p-value of the paired t-test between FedAR and the corresponding baseline (denoted in the columns) for the corresponding dataset (denoted in the rows). From the table, we find that the improvement in performance achieved by FedAR is statistically significant ($p < 0.001$) compared to all the baselines, consistently for both the datasets and both values of $p_{min}$. These results further corroborate the promise and potential of FedAR.
FedAR also shows superior performance on the MNIST dataset with lower training losses and higher test accuracy upon convergence, as elaborated in the Appendix.

\noindent{\textbf{Bias Mitigation.}}
We study the bias mitigation performance of FedAR on CIFAR-10  in the challenging setting, where $p_{min} = 0.1$. Specifically, the global model is used to make predictions for each client after convergence, {and we study the consistency of the prediction accuracies across all clients}. In addition to MIFA and FedVARP, we compare FedAR with the ideal situation of FedAvg, where all the clients are continuous available throughout the entire training process.

\begin{table}[H]
	\centering
	\begin{tabular}{l|cccc}
		\hline
		\small
		ALGO                                                       &
		\begin{tabular}[c]{@{}c@{}}Mean\\ (\%)\end{tabular}        &
		\begin{tabular}[c]{@{}c@{}}Var \\ \\ \end{tabular}         &
		\begin{tabular}[c]{@{}c@{}}Worst 10\% \\ (\%)\end{tabular} &
		\begin{tabular}[c]{@{}c@{}}Best 10\%\\ (\%)\end{tabular}                                                          \\ \hline
		FedAR                                                      & 40.9$\pm$18.1 & 325   & 20.8$\pm$4.5 & 67.7$\pm$8.7  \\
		MIFA                                                       & 34.0$\pm$13.6 & 182.5 & 19.7$\pm$4.2 & 53.8$\pm$8.8  \\
		FedVARP                                                    & 41.3$\pm$20.7 & 432.5 & 19.4$\pm$2.6 & 73.9$\pm$11.6 \\
		FedAvg                                                     & 41.0$\pm$18.0 & 321.6 & 21.2$\pm$4.3 & 69.5$\pm$12.6 \\ \hline
	\end{tabular}
	\caption{Accuracy statistics}\label{client_acc}
	\label{table:accuracy}
\end{table}

\begin{figure}[t]
	\centering
	\includegraphics[width=0.8\linewidth]{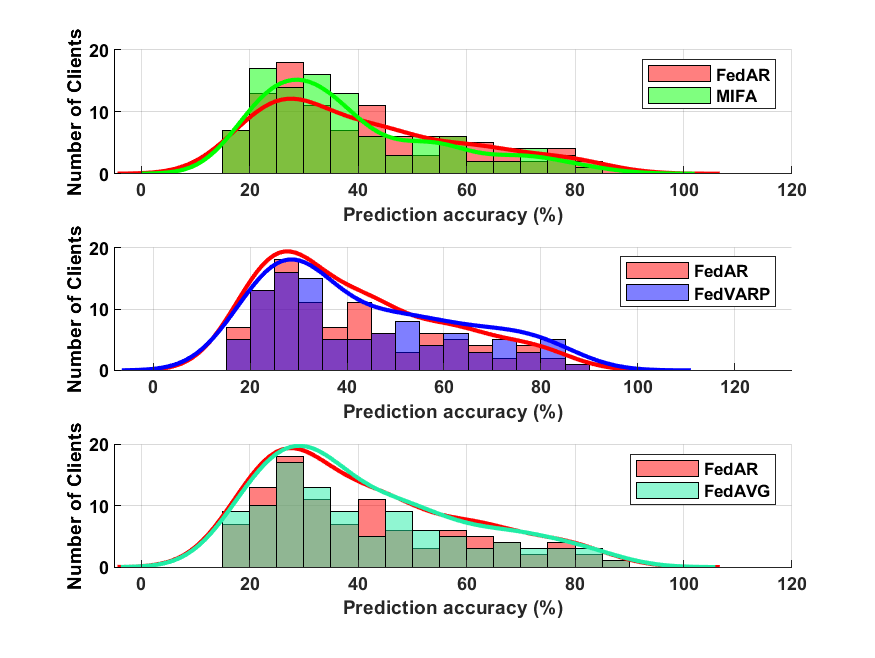}
	\captionof{figure}{Accuracy distributions}\label{fig:bias_p}
	\label{fig:bias}
\end{figure}

Table. \ref{client_acc} depicts the statistics (mean $\pm$ std and variance) of the prediction accuracy across clients. In addition, we record the  prediction accuracy of the worst $10\%$ clients and the best $10\%$ clients, denoted by ``Worst $10\%$'' and ``Best $10\%$'' respectively \cite{li2019fair}. From Table. \ref{client_acc}, we observe that the ``Mean'', ``Worst $10\%$'', and ``Best $10\%$'' prediction accuracy of FedAR  closely align with FedAvg. This suggests that the performance of FedAR is comparable to the ideal situation of full client availability. Furthermore, FedAR achieves an average accuracy approximately $6\%$ higher than MIFA, which requires all the clients to participate in the first training round.  Although the average prediction accuracy of FedVARP is marginally higher than FedAR, it exhibits a considerably higher variance of $432.5$, over $100$ more than FedAR. Such a high variance indicates a significant variation in prediction accuracy across different clients in FedVARP.

To more intuitively evaluate the bias mitigation performance, we visually depict the distribution of the number of clients and its Probability Density Function (PDF) of prediction accuracy in Fig. \ref{fig:bias_p}. Compared to MIFA, FedAR enables a larger number of clients to achieve a prediction accuracy of $40\%$ or higher. Additionally, within the accuracy interval between $25\%$ and $65\%$, the PDF curve of FedAR surpasses that of FedVARP. Outside this interval, PDF curve of FedAR falls below that of FedVARP. This pattern indicates that the prediction accuracies in FedAR are more centralized around the mean value ($40\%$). This {explains the high variance values of FedVARP} in Table. \ref{client_acc}.  Furthermore, the PDF curve of FedAR almost coincides with that of FedAvg.  This indicates that even under the challenging client unavailability ($p_{min}=0.1$), FedAR maintains prediction accuracy distribution similar to the ideal full client availability situation. Both Table. \ref{client_acc} and Fig. \ref{fig:bias_p} confirm that FedAR can effectively mitigate the bias despite severe client unavailability.

\noindent{\textbf{Hyperparameter Evaluation.}}
{We study the effect of hyperparameters under the challenging setting of $p_{min}=0.1$ on CIFAR-10.} Please refer to our Appendix for {the} performance {analysis} on  SVHN and the evaluation for $\rho$ value in Eq. (\ref{compensationscore}).

\begin{figure}[H]
	\centering
	\includegraphics[width=1\linewidth]{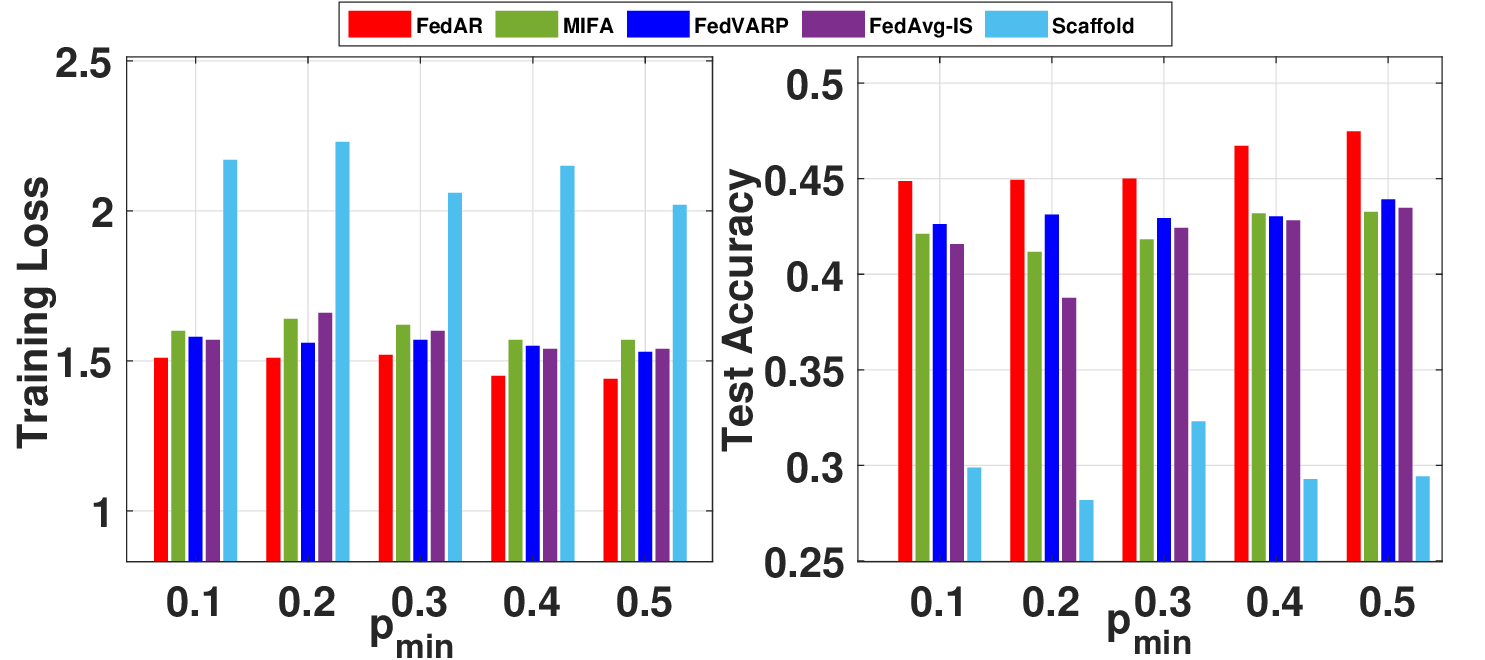}\label{CNN_ALL_P_TL}
	\caption{Effect of $p_{min}$ }\label{CIFARPMIN}
\end{figure}

\noindent\textit{Minimum Client Participation Probability $p_{min}$.} We evaluate FedAR under various client participation probability, i.e., $p_{min}$ spanning from $0.1$ to $0.5$. We exclude FedAvg (S=50) due to its notably inferior performance compared to other baselines. As shown in Fig. \ref{CIFARPMIN}, FedAR consistently outperforms all the baselines for every $p_{min}$. Additionally, we note a marginal enhancement in FedAR's performance as $p_{min}$ increases. When $p_{min}=0.5$, FedAR achieves an accuracy of $47\%$ whereas the accuracy of all the baselines is below $45\%$. This is because a higher participation probability reduces the average number of inactive rounds, thus positively impacting the FL performance.

\begin{figure}[H]
	\centering
	\includegraphics[width=1\linewidth]{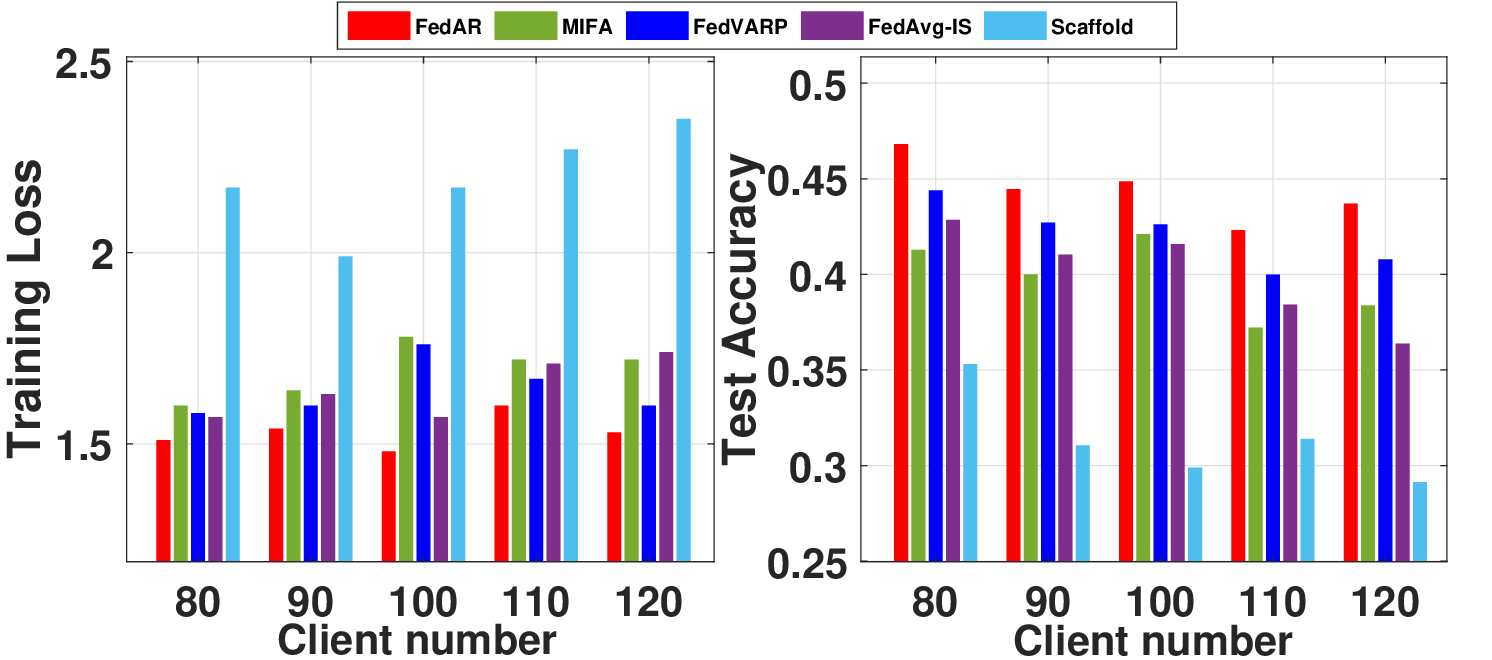}	\label{CNN_ALL_N_TC}
	\caption{Effect of $N$ }\label{CIFAR_N}
\end{figure}

\noindent\textit{{Number of Clients} $N$.} We evaluate FedAR with varying {numbers of clients} from $80$ to $120$. As shown in Fig. \ref{CIFAR_N}, as $N$ increases, all algorithms show an increasing trend in training loss and a decreasing trend in test accuracy, among which FedAR achieves the best performance. Specifically, FedAR marginally increases the training loss only from $1.5$ to $1.6$ when $N$ is increased from $80$ to $120$. The lowest accuracy  of FedAR is $ 43\%$ in the case of $N = 110$. In contrast, the performance of other baselines degrades significantly with the increase in the number of clients.  Except for FedVARP, the test accuracy of the rest of the baselines has fallen below $40\%$. The surge in the number of clients inherently leads to a rise in unavailable clients, posing challenges across all algorithms.  This suggests that FedAR is more adept at handling a large number of clients, making it ideal for large-scale FL, especially in the presence of significant client unavailability.

\section{Conclusion}
In this paper, we propose a novel FL algorithm, FedAR, to address the client unavailability. {We found that}  {clients with different numbers of} inactive rounds have diverse contributions to the current global update. Based on this observation, we design a novel weighting strategy that not only engages the unavailable clients in the global model update, but also {quantifies} their contributions based on the number of their inactive rounds.  We theoretically prove the convergence of FedAR  {for} both convex and non-convex smooth loss functions with non-IID data across clients. Our experimental results demonstrate that FedAR {significantly} outperforms {competing FL baselines} FedAvg, MIFA,  FedVARP and Scaffold {with respect to} the training loss, the test accuracy, and the bias mitigation. FedAR further demonstrates remarkable performance and surpasses those baselines in large-scale FL with severe client unavailability. As part of future work, we will study the performance of FedAR under other practical challenges such as missing data and class imbalance across clients.

\section{Acknowledgment}

The work of X. Zhang is partially supported by the National Science Foundation under Grant Number: CCF-2312617. The work of S.  Chakraborty is partially supported by the National Science Foundation under Grant Number: IIS-2143424 (NSF CAREER Award).

\end{document}